\documentclass[11pt,letterpaper]{article}
\usepackage{graphicx}
\usepackage{naaclhlt2012}
\usepackage{times}
\usepackage{graphics}
\usepackage{subfig}

\title{Extracting Family Relationship Networks from Novels}

\author{
Aibek Makazhanov\thanks{`The author was a student of the 
University of Alberta at the time this research was conducted.}\\
Nazarbayev University Research and Innovation System\\
Astana, Kazakhstan\\
\tt{aibek.makazhanov@nu.edu.kz}
\AND 
Denilson Barbosa \and Grzegorz Kondrak\\
Department of Computing Science\\
University of Alberta\\
 \tt \{denilson,gkondrak\}@ualberta.ca}

\date{}

\begin{document}
\maketitle
\begin{abstract}
We present an approach to the extraction of 
family relations from literary narrative,
which incorporates a technique for utterance attribution 
proposed recently by~\newcite{Elson:2010}.
In our work this technique is used in combination 
with the detection of vocatives ---
the explicit forms of address used by the characters in a novel.
We take advantage of the fact that
certain vocatives indicate family relations between speakers.
The extracted relations are then propagated using a set of rules.
We report the results of the application of 
our method to Jane Austen's {\em Pride and Prejudice}.
\end{abstract}

\section{Introduction}

Relationships between characters can be viewed as both
the basis of the narrative as they become established
and dynamics of it as they evolve.
In this respect family relations 
are naturally established and relatively stable phenomena.
Automated extraction of such relations can assist literary analysis
by providing basic facts for further reasoning on the story.
In the present work we address the problem
of extracting family relations from English narrative.

Relation extraction has been studied in depth over 
the past two decades, and there is a number of existing 
approaches to the problem. 
Among others Open Information 
Extraction approach ~\cite{OpenIE} seems appealing 
due to the unsupervised nature of the method 
and the availability of readily applicable tools. 
We have experimented with one of such tools, 
namely, ReVerb~\cite{rvb}, a state of the art Open IE system. 
Using the tool with the default settings, 
we have extracted a lot of relations from 
Jane Austen's \textit{Pride and Prejudice}.
However, none of those were family relations between two 
characters of the novel.
Apart from giving a rough estimate on the difficulty 
of the addressed problem, this fact also suggests 
that the domain-specific approaches may be of better use.

Traditionally, computer assisted literary analysis
has focused on the word-level methods.
Techniques based on distinguishing word usage 
have been successfully applied
to the tasks of the authorship attribution~\cite{Mostellar},
and the inference of authorial writing style~\cite{Burrows}.
However, an analysis of character interactions
may require a different level of sophistication.
In this respect, techniques for attributing 
utterances to literary characters have been applied to 
conduct an automated evaluation of
different literary hypotheses~\cite{ElsonCvNet}.

Our method extracts family relations from a narrative combining 
word level techniques with utterance attribution approach.
We regard verbal interactions between characters as 
candidate relations, and speakers themselves as their 
potential arguments.
In order to extract the arguments of candidate relations
we attribute all utterances, and 
then apply word-level techniques to extract target relations
from the multitude of candidates.
Specifically, we focus on the usage
of \textit{vocatives} inside the attributed utterances.
Here by a vocative we mean a verbal address in a conversation.
Speakers may address each other by name (named vocative),
or may use phrases such as {\em My dear sister} (nominal vocative).
Nominal vocatives may express sympathy, respect and,
importantly, the family relation between a speaker and 
an intended listener.
We will refer to nominal vocatives simply as vocatives,
and to utterances containing a vocative as vocative utterances.

The intuition behind our approach is to infer family relations 
from nominal vocatives. 
For instance, if in a conversation a speaker 
addresses someone as {\em mother}, 
we can infer a {\em mother of} relation between the speaker  
and an intended listener.
In a dialog scenario such a listener, typically is a speaker of 
the previous or the following utterance. 
We refer to the relations extracted in a described 
manner as seed relations or {\it seeds}.
As the final step our method infers new 
relations from the extracted seeds by applying a 
set of well defined propagation rules, 
e.g. two female characters with a common mother and/or father 
are considered sisters.
Our results suggest that this technique 
trades recall over precision with a positive increase in F-measure.

The structure of the paper is as follows.
We discuss the related work in Section~\ref{RW}.
In Section ~\ref{MTD} we thoroughly describe our method.
Section~\ref{DSD} outlines the properties of our data set.
We present and analyze the results in Section ~\ref{RESS}.
Section~\ref{CONFW} discusses possible directions for future work.

\section{\label{RW}Related Work}

Relation extraction has been studied in depth over considerably long time in different domains and on different scales. 
Most of the effort traditionally was concentrated on developing 
Open RE systems~\cite{OpenIE,rvb}, 
that could easily scale on Web size corpora, and 
required no labeled data or predefined set of relation categories. 
Therefore it is hard to point out works concerned specifically 
with extracting family relations from literary fiction. 
Recently, however, there has been some research in this direction.

\newcite{Santos}~develop a rule based approach towards 
extracting family relations from Portuguese narrative. 
The authors develop 99 rules including some propagation 
rules and combine them with NER and anaphora 
resolution systems. 
The work has been evaluated on two corpora:
bibliographical texts and a collection of sentences.
For both corpora the authors report close results with 
.71 precision and .24 recall under the best settings. 

\newcite{Kokkinakis}~develop an unsupervised approach 
to the extraction of interpersonal relations, including family. 
The method relies on co-occurrences of character names in same 
sentences.
Such co-occurrences are labeled as relations using 
context similarity and on-line lists of common relations. 
The authors evaluate the method on three volumes of Swedish 
XIX century fiction, and report precision in the range of 
91\%-100\% and recall in the range of 70\%-75\% 
under the best settings.
Although the reported results include different types of relations, 
results on the extraction of family relations were not reported.

In contrast to the aforementioned works, our method 
utilizes utterance attribution and vocative detection techniques, 
and employs propagation to infer implicit family relations.

\section{Extracting Family Relations}
\label{MTD}

Our method consists of the following four steps:

\begin{enumerate}
\item Utterance attribution: 
attribute each utterance to one of the characters.
\item Vocatives detection: identify vocative utterances.
\item Relation extraction: extract relations between the speakers
and the characters they address.
\item Relation propagation: derive new relations
between indirectly related characters.
\end{enumerate}

In the following subsections we discuss each of the steps in details.

\subsection{Utterance Attribution}

The cornerstone of our approach is the utterance attribution (UA) method.
In the present work, we employ 
the UA method of~\newcite{Elson:2010}.
Before describing the method we give some basic definitions.
\begin{itemize}
\item {\em Utterance} - A span of narrative enclosed 
in quotation marks. 
Quoted words, phrases, aphorisms, and 
other instances of indirect speech 
enclosed in quotations are left without attribution 
as they are considered {\em non speaker utterances}.

\item {\em Character mention} - a named entity of 
the {\em person} class or a {\em nominal}. 
Nominals consist of a determiner and a head noun. 
Determiners include articles, ordinal and 
cardinal numbers as well as possessives 
({\em her mother}, {\em Elizabeth's sister})\footnote{In 
the original work the list of head nouns is compiled using 
the development corpus. It includes 
nouns that have words {\em person}, {\em man} or 
{\em woman} among the hypernyms of 
the first two WordNet senses ~\cite{wn}.}.

\item {\em Speaker} - a character mention 
to whom certain utterance is attributed.
\end{itemize}

The method is a mixture of the heuristic and supervised approaches: 
based on a category an utterance is either 
immediately attributed, using context
and dialog chain information (heuristic), or 
a feature vector is created per each candidate speaker 
found in the close proximity to an utterance 
and fed to trained classifiers (supervised).
The method attributes each utterance to one of the 
following six categories: 
\begin{enumerate}
\item \textit{Character trigram.} 
Utterances followed or preceded by 
a character mention and an expression verb\footnote{A verb whose
WordNet lexical file is categorized 
as {\em communication} or {\em cognition}.}
(in any order) fall under this category. 
Such an utterance is attributed to a respective character mention. 
\item \textit{Added quote}. 
An utterance that immediately follows another one 
in a paragraph belongs to this category, 
and is attributed to the speaker of the preceding utterance.
\item \textit{Quote alone.} 
An utterance appears by itself in a paragraph. 
The attribution requires a supervised approach.
\item \textit{Apparent conversation.} 
This category is designed to account for dialog chains, 
where two characters speak taking turns. 
To fall under this category an utterance 
must fall under \textit{``Quote alone"} category and 
two previous paragraphs must begin with utterances and 
contain only one utterance each. 
The speaker of the first of the two preceding 
utterances is assigned to the current one.
\item \textit{Anaphora.} 
This category is the same as the 
\textit{``Character trigram"}, but instead of 
character mentions pronouns are treated as possible candidates. 
An utterance is attributed based on a supervised approach. 
Basically, classifiers try to perform anaphora resolution 
given a set of candidate speakers.
\item \textit{Backoff.} 
An utterance is assigned this category if it 
does not fall under any other. 
The attribution requires a supervised approach.
\end{enumerate}

If an utterance cannot be attributed heuristically 
a supervised approach is considered. 
To this end, for each candidate-utterance pair 
a feature vector is extracted. 
Features include various quantitative  
(e.g. a number of appearances of a speaker, a length 
of an utterance, etc.) and qualitative 
(e.g. presence and type of punctuation and verbs 
surrounding an utterance, etc.) characteristics\footnote{A complete 
and thoroughly described list of features 
can be found in the original work~\cite{Elson:2010}.}.
As in the original work, for classification we used 
J48, JRip, and Logistic Regression classifiers as provided by Weka~\cite{weka} and performed a standard 10 fold cross validation.
The predictions generated per each candidate 
are ranked in the following ways:
\begin{enumerate}
\item Label based ranking simply predicts a candidate positively labeled by a classifier. If more than one candidate is labeled positively all candidates are rejected.
\item Single probability based ranking uses probabilities supplied 
by a classifier and predicts the candidate with the highest probability. 
If all probabilities fall below a certain threshold, 
all candidates are rejected.
The threshold is varied as a parameter.
\item Hybrid method first tries the label based ranking and if all candidates are rejected the method employs the probability based ranking.
\item The final ranking method works exactly as the 
single probability method except it combines probabilities 
provided by different classifiers in different 
ways (max, mean, median, product).
\end{enumerate}

\begin{table}[t]
\begin{center}
\begin{tabular}{|c|}
\hline
\emph{mother, father, son, daughter, child,}\\
\emph{sister, brother, cousin, aunt, uncle,}\\
\emph{niece, nephew, wife, husband, grandfather,}\\
\emph{grandmother, mother/father/sister/brother-in-law}\\
\hline
\end{tabular}
\end{center}
\vspace{-5pt}
\caption{A list of target nominals}
\label{tarnomlst}
\end{table}

It is important to note that when attributing speakers to utterances in 
\textit{``Added quote"} and \textit{``Apparent conversation"} 
categories~\newcite{Elson:2010} utilize the ground truth 
information about the speakers of the preceding utterances. 
The authors explain such a behavior as an attempt 
to escape possible propagation errors that would occur 
if preceding utterances were incorrectly attributed. 
We adopt this approach and also rely on the ground truth. 

\subsection{Vocative Detection}

Once we have attributed all utterances,
we need to detect vocatives.
We first define target nominals, 
i.e. nouns that describe family relations.
A list of basic family relations is shown in Table~\ref{tarnomlst}. 
We further extend it by including WordNet synonyms and hypernyms 
for each nominal to have a total of 635 target nominals.
Having the list ready, we proceed to select 
candidate utterances for the vocative detection task. 
An utterance is considered a candidate 
if it contains at least one target nominal.

For the detection task we tried both 
unsupervised and supervised methods. 
The unsupervised method predicts a vocative 
utterance if a candidate matches the following pattern: 
$<$P {\em my dear(est)} T P$>$. 
Here T stands for the target nominal 
and P denotes punctuation marks.
We consider all punctuation marks 
commonly used, including quotations.
The occurrence of words {\em my} and {\em dear(est)} is optional.

Our supervised method is the extended version 
of the approach proposed by ~\newcite{huasthesis}.
The method extracts a feature vector per each 
occurrence of a target nominal in a given candidate utterance.
We use the following set of binary features:

\begin{itemize}
\item \textit{Lexical features.} 
Whether words \emph{``my"} and \emph{``dear(est)"} precede 
a nominal either by themselves or in combination. 
Whether a nominal is preceded or followed 
(ignoring punctuation) by the word \emph{``you"}. 
Whether the word \emph{``oh"} appears anywhere 
before a nominal in an utterance. 
Whether the words {\em ``you"} or {\em ``my dear"} appear in an 
utterance regardless of the distance and direction towards a nominal.
\item \textit{Punctuation.} Whether a comma, period, question or exclamation marks appear immediately to the right (ignoring spaces) of a nominal. Whether a nominal is surrounded by commas or by any other punctuation marks, including quotations.
\item \textit{Positional and other features.} Whether a nominal is 
found in the beginning or the end of an utterance. 
Whether a nominal occurs more than once in an utterance. 
\end{itemize}

To generate the predictions we manually label extracted 
feature vectors and perform a standard 10 fold cross validation.

\subsection{Relation Extraction}

Once we have identified vocative utterances 
we proceed with the relation extraction.
We extract relations in the form of 
a standard (A1, R, A2) triple, where A1 and A2 
denote the arguments of a relation, 
and R denotes the relation itself. 
From a given vocative utterance we can immediately recover the 
relation, which is the vocative itself, e.g. {\em mother}, {\em son}, etc. 
Moreover, the speaker assigned to the utterance becomes 
the second argument of the relation. 
So, the main challenge is to extract the first argument of the relation, 
or, in other words, is to identify a recipient to whom a vocative was addressed.

Given an utterance we have up to two potential recipients of the 
vocative: the speakers of the preceding and the following utterances. 
Sometimes one or even both may not be present. 
To eliminate unsuitable candidates 
we introduce the following three constrains:

\begin{enumerate}
\item The gender of a candidate must match 
the gender of a vocative, 
i.e. if a vocative is \emph{``my dear sister"} we reject male candidates.
\item The utterance spoken by a candidate 
must be in the paragraph immediately 
preceding or following the given vocative utterance.
\item If both candidates satisfy the first two 
constrains the speaker of the following utterance is chosen.
\end{enumerate}

The last two constrains were set empirically based on 
the experiments with a development corpus. 
If both candidates got rejected, we abandon a relation.

\begin{table}[!t]
\begin{center}
\begin{tabular}{|l|}
\hline
\emph{\bf \centering Compliment rules}\\
\hline
 1. (A, \emph{cousin of}, B) $\Rightarrow$ (B, \emph{cousin of}, A)\\
 2. (A, \emph{wife / husband of}, B) $\Rightarrow$ \\(B, \emph{husband / wife of}, A)\\
 3. (A, \emph{Mr. and Mrs.}, B) \& \\FEMALE(B) $\Rightarrow$ (B, \emph{wife of}, A)\\
\hline
\emph{\bf \centering Transitivity rules}\\
\hline
1. (A, \emph{cousin of}, B) \&\\
(B, \emph{cousin of}, C) $\Rightarrow$ (A, \emph{cousin of}, C)\\
2. (A, \emph{sister / brother of}, B) \&\\
(B, \emph{sister / brother of}, C)
$\Rightarrow$ \\ (A, \emph{sister / brother of}, C)\\
\hline
\emph{\bf \centering Compound rules}\\
\hline
1. (A, \emph{father / mother of}, B) \&\\
(B, \emph{sister / brother of}, C) $\Rightarrow$ \\(A, \emph{father / mother of}, C)\\
2. (A, \emph{father / mother of}, B) \&\\
(C, \emph{sister / brother of}, A) $\Rightarrow$ \\(C, \emph{aunt / uncle of}, B)\\
\hline
\end{tabular}
\end{center}
\vspace{-5pt}
\caption{\label{proprules} Sample propagation rules by categories}
\end{table}

\subsection{Relation Propagation}

As the final step we try to infer new relations from the extracted 
seed relations using a rule based propagation technique. 
We have applied a total of 21 rules to further propagate the seeds. 
These rules can be divided into three major categories:

\begin{itemize}
\item Simple compliment rules
\item Transitivity rules
\item Compound rules
\end{itemize}

Table~\ref{proprules} lists sample propagation rules by categories. 
Each rule has an analogue that works in the opposite direction. 
Similarly each gendered rule has an analogue 
that works for the opposite gender.
Depending on the accuracy of an initial extraction propagation 
rules may arrive to contradictions. 
For example, suppose we have 
extracted the following relations: (i) (A, \emph{father of}, B), 
(ii) (C, \emph{sister of}, A) and (iii) (C, \emph{sister of}, B). 
According to the second compound rule in Table~\ref{proprules} 
we can propagate the relation (C, \emph{aunt of}, B) 
from the first two relations. 
However, it contradicts to the already existing third relation.
We solve possible contradictions with the help of relation counts. 
Let us assume that the first relation from the previous 
example were found $n$ times during the extraction, and 
the second and the third relations were found $m$ and $q$ 
times respectively. 
We propagate these counts together with the relations. 
If the maximum of counts corresponding to the relations on the 
left hand side of a rule is larger than the count of 
the existing contradictory relation, we replace such a relation 
with the propagated one, otherwise we cancel the propagation. 
In the case of our example we propagate 
the (C, \emph{aunt of}, B) relation if $max(n,m)>q$.

\begin{table}[t]
\begin{center}
\begin{tabular}{l|r|r}
\hline & \bf Dev. corpus & \bf Test corpus \\ \hline
chapters & 23 & 38\\
utterances & 734 & 1019\\
utterances with & &\\
target nominals & 67 & 251\\
vocative utterances & 10 & 38\\
\hline
\end{tabular}
\end{center}
\vspace{-5pt}
\caption{\label{data-des} Data set description }
\vspace{-10pt}
\end{table}

\section{\label{DSD}Data Set Description}

We have developed and tested our method on 
Jane Austen's {\em Pride and Prejudice}. 
A fairly high number of characters and a rich set 
of family relations make this novel a plausible choice. 
Moreover, in Victorian novels family relations are 
well defined, making propagation rules perfectly applicable.
We used first volume of 23 chapters as the 
development corpus and keep the rest for the evaluation.
Basic characteristics of the data 
are given in Table~\ref{data-des}.
We will refer to the test set as the PnP corpus.

We have annotated the novel in the Columbia 
QSA corpus fashion  \cite{Elson:2010}, 
labeling named entities and nominals exactly as 
they were labeled in the original work.
The ground truth information obtained from human annotators 
includes utterance attribution, 
a list of family relations and a gender-labeled list of characters.

The list of family relations has a total of 202 relations 
between 28 characters. 
All relations are bidirectional.
The gendered list of characters also includes 
cross-linked references for character names, i.e. mentions like 
Elizabeth, Miss Eliza, Eliza, etc. are linked as the same entity.
Although we did perform gender attribution 
and co-reference resolution for NEs, in the present 
work we use the ground truth information.

\section{\label{RESS}Results}

In this section we report the performance of our 
method at each step that contributes to the final result. 
We start with the utterance attribution and 
compare our implementation of the method 
to the original work.
We then report results for the vocative detection task. Here we compare supervised and unsupervised approaches.
Finally, we report results of the relation extraction, 
assess the impact of the propagation step, and 
carry out the error analysis.

\begin{table}
\begin{center}
\begin{tabular}{l|l|r|r}
\hline \bf Method & \bf Corpus & \bf \# utterances & \bf Accuracy \\
\hline
 Original & CQSA & 3064 & .83\\
\hline
 Ours & CQSA & 3064 & .78\\
 Ours & PnP & 1014 & .79\\
\hline
\end{tabular}
\end{center}
\caption{\label{elRes} Performance of our implementation of QSA method on different corpora compared to the original work }
\end{table}

\subsection{Utterance Attribution}

Table~\ref{elRes} shows the performance of utterance attribution method on different corpora. 
In order to assess the quality of our implementation of 
the method by~\newcite{Elson:2010}, 
we report the overall accuracy of our method 
on the data used in the original work, i.e. 
Columbia Quoted Speech Attribution (CQSA) corpus.
In this experiment our implementation achieves the accuracy of 78\% which is 5\% lower than the results 
reported in the original work.
We found that heuristics perform closely to the original method, 
but the classifiers do not perform as good. 
On the PnP corpus the overall accuracy of the method was 79\%. 
A closer analysis of the results revealed 
a trend similar to the one observed for the CQSA corpus: 
heuristics performed better than classifiers.

\begin{table}
\begin{center}
\begin{tabular}{l|r|r|r}
\hline \bf Method & \bf P & \bf R & \bf F \\ \hline
Naive Bayes & .90 & .87 & .88\\
Pattern matching baseline & .47 & .92 & .62\\
\hline
\end{tabular}
\end{center}
\caption{\label{vocpred} Results of the vocative prediction task}
\end{table}

\subsection{Vocative Detection}

Table~\ref{vocpred} shows the results of the vocative detection task. 
When experimenting with our supervised method 
we tried a number of classifiers.
The Naive Bayes as provided by Weka achieved the 
highest F-measure, and recognized 33 vocative utterances, making 
four false positive predictions.
The unsupervised method, however, despite achieving higher recall, 
produced twice as much false positives as true positives. 
This suggests that pattern matching, 
although unsupervised, can potentially lead to 
a high false positives rate at the relation extraction step. 
Therefore, at the extraction step we use vocative utterances 
detected by the supervised method.

\subsection{Relation Extraction and Propagation}

Both relation extraction and propagation 
depend on the accuracy of the utterance attribution. 
In order to account for this dependency and for 
comparison purposes we introduce the oracle, 
which utilizes ground truth speaker information 
at the relation extraction step; 
hence, it is free from the utterance attribution errors.
Also, we would like to assess the impact 
of propagation from both positive and negative angles. 
Propagation increases recall, but 
due to possible propagation errors it may hurt precision. 
In order to account for such errors 
we employ manual cleaning of the 
seed relations after the extraction step.
Cleaning entails removing the incorrect seeds, not correcting them.

\begin{figure}[!t]
\begin{minipage}[t]{230pt}
 \subfloat[Extracted seeds]
{\label{rels_pd}\includegraphics[width=\textwidth]{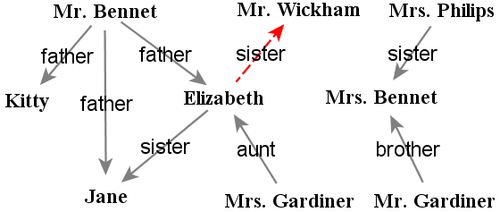}}
\vspace{14pt}
  \end{minipage}
\begin{minipage}[t]{230pt}
 \subfloat[Propagated cleaned seeds]{\label{rels_pdpc}\includegraphics[width=\textwidth]{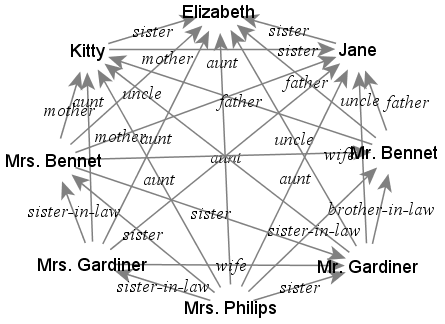}}
  \end{minipage}
  \caption{Extracted seed relations before (a) and 
after (b) cleaning and propagation. Dashed arrows 
indicate inaccurate extractions.} 
  \label{rels}
\vspace{-10pt}
\end{figure}

\begin{figure}[!t]
\begin{minipage}[t]{230pt}
 \subfloat[Oracle seeds]{\label{rels_gt}\includegraphics[width=\textwidth]{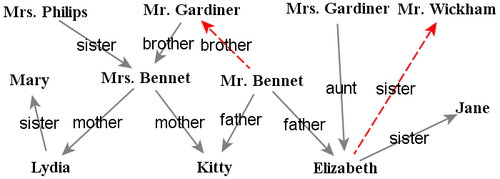}}
  \end{minipage}
\begin{minipage}[t]{230pt}
 \subfloat[Propagated cleaned oracle seeds]{\label{rels_gtpc}\includegraphics[width=\textwidth]{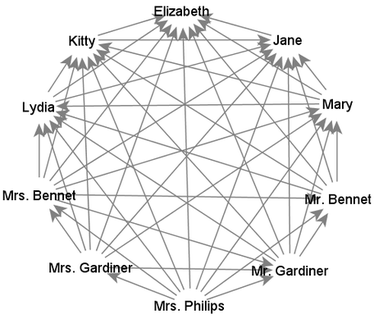}}
  \end{minipage}
  \caption{Oracle seed relations before (a) and 
after (b) cleaning and propagation. Dashed arrows 
indicate inaccurate extraction.} 
  \label{rels2}
\vspace{-10pt}
\end{figure}

Table~\ref{basic_rex} shows  
precision/recall of the relation extraction step. 
Precision of the extraction is fairly high, 
in fact it, is higher than that of the oracle. 
This happens due to semantic errors which we 
will discuss later. 
Both extracted and oracle seeds have a 
low recall, which is depicted by the 
sparsity of networks shown 
on Figures ~\ref{rels_pd} and ~\ref{rels_gt}. 
Cleaning obviously results in a 100\% 
precision, but it has no effect 
on recall at this stage.

\begin{table}[!t]
\begin{center}
\begin{tabular}{l|c|c|c}
\hline
& \bf P & \bf R & \bf F \\
\hline
Extracted seeds
& \raggedright 0.88 & 0.03 & 0.06\\
Cleaned seeds
& \raggedright 1.00 & 0.03 & 0.06\\
Oracle seeds
& \raggedright 0.82 & 0.04 & 0.08\\
Cleaned oracle seeds
& \raggedright 1.00 & 0.04 & 0.08\\
\hline
\end{tabular}
\end{center}
\caption{\label{basic_rex} Results of the 
relation extraction task}
\end{table}

Table~\ref{clean_rex} shows results of 
propagating all four sets of seed relations listed in 
Table~\ref{basic_rex}. 
The first line of Table~\ref{clean_rex} shows the 
final, overall performance of our method. 
Precision of the method was 77\%, 
which is very close to the oracle results.
However, 27\% recall is still 
lower than the oracle performance.
As we have expected propagation have increased 
recall and slightly decreased precision. 
Propagation resulted in 9 and 11 times 
increase in recall for extracted and oracle 
seeds respectively. 
The respective decrease in precision was 
1.14 and 1.05 times.
In both cases positive increase in F-measure is 
obvious, which speaks in favor of propagation.
Cleaning does not seem to have 
notable effect on recall, as propagating extracted and oracle seeds 
with and without cleaning yields almost the same recall. 
However, the main advantage of cleaning is that it yields a 100\% 
precision and no propagation errors.

\begin{table}[!t]
\begin{center}
\begin{tabular}{l|c|c|c}
\hline
& \bf P & \bf R & \bf F \\
\hline
Extracted seeds
& \raggedright 0.77 & 0.27 & 0.40 \\
Cleaned seeds
& \raggedright 1.00 & 0.27 & 0.42\\
Oracle seeds
& \raggedright 0.78 & 0.43 & 0.55\\
Cleaned oracle seeds
& \raggedright 1.00 & 0.44 & 0.61\\
\hline
\end{tabular}
\end{center}
\caption{\label{clean_rex} Results of the propagation task}
\end{table}

Figures ~\ref{rels_pdpc} and ~\ref{rels_gtpc} 
show extracted and oracle seeds after the cleaning 
and propagation. For the clarity of presentation 
we do not show duplicate edges. Also, for the 
same reasons, we drop relation labels on the latter 
figure. 
The increase in recall is apparent from the increase 
in densities of the respective networks.

Lastly we would like to point out the importance of 
one particular propagation rule. 
Clearly, the network on Figure~\ref{rels_pd} is disconnected, 
but after propagation it becomes connected as shown on 
Figure~\ref{rels_pdpc}. 
This happens due to the 3rd compliment rule listed in 
Table~\ref{proprules}. 
According to the rule we infer {\em wife of} relation 
between Mrs. Bennet and Mr. Bennet and Mrs. Gardiner and 
Mr. Gardiner on the basis of the titles and 
the same last names. 
This connects the network of extracted seeds and 
makes further propagation possible.

\subsection{Error Analysis}

In this section we discuss possible problems with our approach. 
First, we would like to point out the impact of 
inaccurate utterance attribution. 
Comparing Figures ~\ref{rels_pd} and ~\ref{rels_gt} one can 
immediately notice that the former one misses 
the nodes {\em Mary} and {\em Lydia} and the relations 
(Lydia, {\em sister of}, Mary) and 
(Mrs. Bennet, {\em mother of}, Lydia).
This is the main reason why recall of 
the propagation of extracted seeds was 
lower than that of the oracle seeds. 
In the case with the relation between Lydia and Mary, 
a vocative utterance was mis-attributed to Lydia, 
resulting in an invalid (Lydia, {\em sister of}, Lydia) relation, 
which was rejected. 
Similarly, an utterance was mis-attributed 
to Lydia again, in the case of a relation between 
Lydia and Mrs. Bennet. 
This also resulted in a rejected relation.
%
In short, for our method it really matters 
to whom a given utterance was mis-attributed to.

Unfortunately, there is another problem 
with some vocative utterances. 
Notice dashed, incorrect arrows in Figures ~\ref{rels_pd} 
and ~\ref{rels_gt}. 
(Elizabeth, \emph{sister of}, Mr. Whickham) relation was 
extracted because Whickham addressed Elizabeth as sister 
meaning sister-in-law. 
Similarly Mr. Gardiner addressed Mr. Bennet as 
brother meaning brother-in-law which 
lead to inaccurate extraction. 
In fact, these were the only extraction errors 
our method produced. 
This problem is of semantic nature and 
we will seek ways to solve it in our future work.

\section{\label{CONFW}Conclusion and Future Work}

We have developed a method that 
extracts family relations from a narrative
based on the utterance attribution and 
the vocative detection techniques.
Our method successfully applies propagation rules to 
infer new relations from initially extracted ones.
We carried out an error analysis and concluded 
that the method is sensitive to 
the accuracy of utterance attribution and 
vulnerable to the semantics of vocatives. 
We plan to work on these issues in the future.
As the additional line of work 
we plan to explore context clues. For example, 
given a passage like \emph{"Mr. Bennet," replied his wife}, 
we can infer (current speaker, \emph{wife of}, Mr. Bennet) 
relation from the utterance and the surrounding context.

\bibliographystyle{naaclhlt2012}
\bibliography{main}

\end{document}